\documentclass[runningheads]{llncs}
\usepackage{amsmath}
\usepackage{amsfonts} 
\usepackage{graphicx}
\usepackage{ntheorem}
\usepackage{multirow}

\usepackage{color}
\definecolor{ToDoColor}{rgb}{0.1,0.2,1}
\definecolor{CommentColor}{rgb}{0.2,0.8,0.2}

\theoremseparator{:}
\newtheorem{hyp}{Hypothesis}
\newtheorem{examp}{Example}
\newtheorem{clm}{Claim}

\newcommand{\burst}{\texttt{burst}}
\newcommand{\scd}{\texttt{sc}}
\newcommand{\mcd}{\texttt{mc}}

\newcommand{\albd}{\texttt{aLBD}}
\newcommand{\albdsc}{\texttt{aLBD$_\mathtt{sc}$}}
\newcommand{\albdmc}{\texttt{aLBD$_\mathtt{mc}$}}

\newcommand{\aburst}{\texttt{avgBurst}}
\newcommand{\mburst}{\texttt{maxBurst}}

\newcommand{\sumlbd}{\mathtt{sumLBD}}

\newcommand{\minalbd}{\texttt{min\_LBD}}

\newcommand{\fdsc}{\texttt{PDSC}}
\newcommand{\fdmc}{\texttt{PDMC}}

\newcommand{\avgminalbd}{\texttt{avg\_min\_LBD$_\texttt{mc}$}}

\newcommand{\freq}{\texttt{count}}

\newcommand{\mpldl}{\texttt{MplDL}}
\newcommand{\mpldlext}{\texttt{MplDL$^\mathtt{crvr}$}}
\newcommand{\kissat}{\texttt{Kissat-sat}}
\newcommand{\kissatext}{\texttt{Kissat-sat$^\mathtt{crvr}$}}
\newcommand{\kissatdef}{\texttt{Kissat-default}}

\newcommand{\kissatdefext}{\texttt{Kissat-default$^\mathtt{crvr}$}}

\newcommand{\proffac}{\texttt{cp}}

\newcommand{\satcomp}{SAT20}

\newcommand{\trackcrv}{\texttt{DetectPoorCRV}}

\newcommand{\reducecrv}{\texttt{CRVRBranching}}

\newcommand{\poortilted}{\texttt{poor}}

\newcommand{\block}{\texttt{block}}

\newcommand{\lbprc}{\mathbf{LBP_{\mathcal{R}_{\mathcal{C}}}}}

\newcommand{\pfrc}{\mathtt{cp_{\mathcal{R}_{\mathcal{C}}}}}

\newcommand{\rc}{\mathcal{R}_{\mathcal{C}}}

\newcommand{\urc}{U_{\mathcal{R}_{\mathcal{C}}}}

\newcommand{\confcloseness}{\texttt{ConflictsProximity}}

\newcommand{\proximated}{\textit{closely related}}
 
\newcommand{\ctodl}{\mathcal{D}}

\newcommand{\excmpldl}{\mathtt{CRVR\!\!-\!\!bad}}

\newcommand{\excmpldlext}{\mathtt{CRVR\!\!-\!\!good}}

\newcommand{\crvr}{\texttt{CRVR}}

\begin{document}
\title{A Deep Dive into Conflict Generating Decisions}

\author{
Md Solimul Chowdhury,
Martin M{\"{u}}ller,
Jia-Huai You
}
\institute{Department of Computing Science,
University of Alberta.
\\
\email{\{mdsolimu, mmueller, jyou\}@ualberta.ca}}

\maketitle              %

\begin{abstract}Boolean Satisfiability (SAT) is a well-known NP-complete problem. Despite this theoretical hardness, SAT solvers based on Conflict Driven Clause Learning (CDCL) can solve large SAT instances from many important domains. CDCL learns clauses from conflicts, a technique that allows a solver to prune its search space. The selection heuristics in CDCL prioritize variables that are involved in recent conflicts. While only a fraction of decisions generate any conflicts, many generate multiple conflicts.

In this paper, we study conflict-generating decisions in CDCL in detail. We investigate the impact of single conflict (sc) decisions, which generate only one conflict, and multi-conflict (mc) decisions which generate two or more. We empirically characterize these two types of decisions based on the quality of the learned clauses produced by each type of decision. We also show an important connection between consecutive clauses learned within the same mc decision, where one learned clause triggers the learning of the next one forming a chain of clauses. This leads to the consideration of similarity between conflicts, for which we formulate the notion of conflictsproximity as a similarity measure. We show that conflicts in mc decisions are more closely related than consecutive conflicts generated from sc decisions. Finally, we develop Common Reason Variable Reduction (CRVR) as a new decision strategy that reduces the selection priority of some variables from the learned clauses of mc decisions. Our empirical evaluation of CRVR implemented in three leading solvers demonstrates performance gains in benchmarks from the main track of SAT Competition-2020.

\end{abstract}

%
%
%
\section{Introduction}
Boolean Satisfiability (SAT) is a fundamental problem in computer science, with strong relations to computational complexity, logic, and artificial intelligence. Given a formula $\mathcal F$ over boolean variables, a SAT solver either determines a variable assignment which satisfies $\mathcal F$, or reports unsatisfiability if no such assignment exists. In general, SAT solving is intractable \cite{Cook71}. Complete SAT solvers based on the 
 framework of DPLL \cite{DPLL} employ heuristics-guided backtracking tree search.
CDCL SAT solvers such as GRASP \cite{GRASP} and Chaff \cite{Chaff}
substantially enhanced the DPLL framework by adding 
conflict analysis and clause learning. 

Modern CDCL SAT solvers can solve very large real-world problem instances from important domains such as hardware design verification \cite{HardwareDesign}, software testing \cite{exeDebugger},  automated planning \cite{satPlanning}, and encryption \cite{logicalCrypt,cryptoMiniSAT}. The efficiency of modern solvers is the result of careful integration of their key components such as preprocessing \cite{preprocessing,BCE}, inprocessing \cite{inprocessing,LCM}, robust decision heuristics \cite{LiangGPCAAAI16,LiangGPCSAT16,Chaff}, efficient restart policies  \cite{glucoseRestart,BetweenSATUnsat}, intelligent conflict analysis \cite{GRASP}, and effective clause learning \cite{Chaff}. 

Clauses learned in CDCL can help prune the search space. As finding conflicts is the only known efficient way to learn clauses, the rate of conflict generation is critical for CDCL SAT solvers. State-of-the-art decision heuristics such as Variable State Independent Decaying Sum (VSIDS) \cite{Chaff} and Learning Rate Based (LRB) \cite{LiangGPCSAT16} prioritize the selection of variables which appear in recent conflicts. 

It was shown empirically \cite{Liang2017} that the most efficient CDCL decision heuristics produce about 0.5 conflicts per decision on average. The clause learning process in CDCL can generate more than one conflict for one decision. In the following, we categorize each conflict-producing decision as a \textit{single conflict} (\scd{}) or a \textit{multi-conflicts} (\mcd{}) decision, depending on whether it produces one, or more than one, conflict.  We label the resulting learned clauses \scd{} and \mcd{} clauses accordingly.

 Conflicts play a crucial role in CDCL search. A better understanding of conflict generating decisions is a step towards a better understanding of CDCL and may open up new directions to improve CDCL search. Motivated by this, here 
we study conflict producing decisions in CDCL. The contributions of this work are: 
\begin{itemize} 
\item [1.] We compare \scd{} and \mcd{} decisions in terms of the average quality of the learned clauses. It turns out that the average LBD score is significantly lower (of better quality) for \scd{} than for \mcd{} clauses. 

\item [2.] We analyze the distribution of 
conflicts in \mcd{} decisions, which shows that although a \mcd{} decision can produce a large number of consecutive conflicts, \mcd{} decisions with a low number of consecutive conflicts are more frequent.


\item [3.] An analysis that shows how consecutive clauses learned by a \mcd{} decision are connected to each other. 

\item [4.]
We introduce the measure of \confcloseness{} to study the relation between conflicts in a given conflict sequence. The proximity between a set of conflicts is defined in terms of literal blocks that are shared between the reason clauses of these conflicts.  
We show that the conflicts which are discovered during the same \mcd{} decisions are closer by this measure than conflicts discovered by consecutive \scd{} decisions.

\item [5.] We develop a CDCL decision strategy, called \textit{Common Reason Variable Reduction} (\crvr{}), which reduces the priority of some variables that appear in \mcd{} clauses. Our empirical evaluation of \crvr{} on benchmarks from the maintrack of SAT Competition-2020 shows performance gains 
for satisfiable instances in several leading solvers.
\end{itemize}
\section{Preliminaries}
\label{sect_prelim}
\subsection{Background}
\subsubsection{Basic Operations of CDCL} 
Given a boolean formula $\mathcal F$,
a CDCL SAT solver works by extending an (initially empty) {\em partial assignment}, 
a set of literals representing how the corresponding variables are assigned. In each \textit{branching decision} (or just {\em decision}),
the solver extends the current partial assignment by selecting a single
 variable $v$, called a {\em decision variable},  from the current set of unassigned variables, 
 and assigns a boolean value to it.
A decision is associated with a \textit{decision level} 
$\mathit{dl} \ge 1$: the depth of the search tree when the decision was made. 
Then, 
\textit{unit propagation} (UP) is invoked to 
simplify $\mathcal{F}$ by deducing a new set of implied variable assignments, which are 
added to the current partial assignment. After UP, 
if $\mathcal{F}$ is still unsolved and no conflict occurs,
the search moves on to the next decision level and the process repeats.

\medskip
\noindent
{\bf Conflicts and Clause Learning} UP may lead to a \textit{conflict}
due to a  \textit{conflicting} clause $C$, which cannot be satisfied under the current partial assignment. In this case, \textit{conflict analysis} generates a learned clause and a backtracking level. The learned clause can help to prune the remaining search space.  

Most state-of-the-art CDCL SAT solvers employ the \textit{first Unique Implication Point (fUIP)} scheme to learn a clause. Starting with conflicting clause $C$, fUIP continues to resolve literals from the current decision level until it finds a clause $L=R \lor \{\lnot f\}$ such that the literal $f$ was assigned in the current decision level, while all literals in $R$ were assigned earlier. $f$ is called the {\em fUIP literal} for the current conflict and is contained in every path from the current decision variable to the current conflict. 
The literals in $\{r\; \vert \;\textnormal{literal}\; r \in R\} \cup \{f\}$ are called \textit{reason literals} for the current conflict, since their assignments caused the current conflict. We call $R$ the \textit{reason clause} for the current conflict with conflicting clause $C$.

After $L$ is learned, search backjumps to a \textit{backjumping level} \texttt{bl} which is computed from $L$\footnote{If \texttt{bl} is too far from the current decision level, then performing chronological backtracking results in better solving efficiency \cite{chronobt}. Most of the leading CDCL solvers employ a combination of chronological and non-chronological backtracking.}. Before backtracking, the clause $L=R \lor \lnot f$ is unsatisfied under the current partial assignment. After backtracking to \texttt{bl}, $\lnot f$ is the only unassigned literal in $L$. The search proceeds by unit-propagating $\lnot f$ from $L$. The assignment of $\lnot f$ avoids the conflict at $C$, but may create further conflicts within the same decision, making this a \mcd{} decision.


	
	
	
\medskip
\noindent
{\bf Relevant Notions} 
The following notions have been proposed in the literature, which are relevant for our paper:
\hfill\\
\textbf{Global Learning Rate:} The Global Learning Rate (GLR) \cite{Liang2017} is defined as $\frac{c}{d}$, where $c$ is the number of conflicts generated in $d$ decisions.  GLR measures the average number of conflict per decisions of a solver.

 \noindent\textbf{The Literal Block Distance  (LBD) Score:}  
The LBD score \cite{AudemardS09} of a learned clause $c$ is the number of distinct decision levels in $c$. If LBD$(c)\!=\!n$, then $c$ contains $n$ propagation blocks, where each block has been propagated within the same decision level. Intuitively, variables in a block are closely related. Learned clauses with lower LBD score tend to have higher quality.

\noindent \textbf{Glue Clauses}  \textit{Glue clauses} \cite{AudemardS09} have LBD score of 2 and are the most important type of learned clauses.  A glue clause connects a literal from the current decision level with a block of literals assigned in a previous decision level. Glue clauses have the strongest potential to reduce the search space quickly. 

\noindent\textbf{Glue to Learned (G2L)}. This measure represents the fraction of learned clauses that are glue clauses \cite{CP19}. It is defined as $\frac{g}{c}$, if there are \textit{g} glue clauses among \textit{c} learned clauses.

\subsection{Notation}
We denote a CDCL solver $\psi$ running a given SAT instance $F$ by $\psi_F$. Assume that this run makes $d$ decisions and generates $c$ conflicts.
	
\medskip
\noindent
{\bf \scd{} and \mcd{} Decisions}
A \scd{} decision generates exactly one conflict and learns a \scd{} clause,
while a \mcd{} decision generates more than one conflict and accordingly learns multiple \mcd{} clauses. 
Let $\psi_F$ takes $s$ \scd{} decisions and $m$ \mcd{} decisions, learning $c_s$ and $c_m$ clauses respectively. Then $d=m+s$ and $c=c_s+c_m$.


\medskip
\noindent
{\bf Burst of \mcd{} Decisions}
We define the \burst{} of a \mcd{} decision as the number of conflicts (i.e., learned clauses) generated within that \mcd{} decision. 

For $\psi_F$, we define
\begin{itemize}
\item \aburst{}, the average burst over $m$ \mcd{} decisions as  $\frac{c_m}{m}$
    \item \mburst{}, the maximum burst among the bursts of $m$  \mcd{} decisions.  
    \item  We define a mapping $\freq_b: \mathbb{Z}  \longmapsto \mathbb{Z}$ which takes a burst $b \ge 2$ as input and outputs the number of \mcd{} decisions with burst $b$. 
\end{itemize}

\medskip
\noindent
{\bf Learned Clause Quality Over \scd{} and \mcd{} Decisions}
Let $lbd(L)$ be the LBD score of a learned clause $L$.  For $\psi_F$, we define
\begin{itemize}
    \item the average LBD score \albd{} over $c$ learned clauses 
    as $\frac{\sumlbd}{c}$, where $\sumlbd{}$ is the sum of LBD scores over $c$ learned clauses. 
    \item  the average LBD score \albdsc{} over $c_s$ \scd{} clauses as $\frac{\sumlbd_{\scd}}{c_s}$, where $\sumlbd_{\scd}{}$ is the sum of LBD scores over $c_s$  clauses. 
     \item the average LBD score \albdmc{} over $c_m$ \mcd{} clasues as $\frac{\sumlbd_{\mcd}}{c_s}$, where $\sumlbd_{\mcd}{}$ is the sum of LBD scores over $c_m$ clauses.
\end{itemize}  
For \mcd{} decision $\mathcal{M}$, we denote the minimum LBD score among its learned clauses by $\minalbd_{\mathcal{M}}$. For $\psi_F$, \avgminalbd{} is the average minimum LBD over $m$ \mcd{} decisions.

\subsection{Test Set, Experimental Setup and Solvers Used}\label{test-set}
All our experiments use the following setup:
The test set consists of 
400 benchmark instances from the main-track of SAT Competition-2020 (short \satcomp{}) \cite{sc20downloads}.
The timeout is 5,000 seconds per instance.
Experiments were run on a Linux workstation with 64 Gigabytes of RAM, processor clock speed of 2.4GHZ, with L2 and L3 caches of size 256K and 20480K, respectively.

We use the following solvers for evaluation:
in Section \ref{section_mc_sc_study}, we
use the solver MapleLCMDiscChronoBT-DL-v3 \cite{mpldl} (short \mpldl{}), 
the winner of SAT Race-2019, and 
in Section \ref{crvr}, we extend three
leading CDCL SAT solvers: \mpldl{} and the top two solvers in the main track of SAT Competition-2020, Kissat-sc2020-sat 
(\kissat{}) and Kissat-sc2020-default  (\kissatdef{}) \cite{sc20downloads}.

\section{An Empirical Analysis of \scd{} and \mcd{} Decisions}
\label{section_mc_sc_study}
In this section, we present our empirical study of conflict-generating decisions in CDCL search.
We use
\mpldl{} as CDCL solver and investigate its \scd{} and \mcd{} decisions. 
\begin{table}[h]
\centering
\caption{Conflict Generating Decisions. Columns A to G shows average measures over the number of instances shown in the Count column.}
\label{TabCGD}
\resizebox{\columnwidth}{!}{
\begin{tabular}{|c|c||c|c||c|c|c||c|c|}
\hline
\multirow{2}{*}{\textbf{Type}} & \multirow{2}{*}{\textbf{Count}} & \multicolumn{2}{c||}{\textbf{Conflict Frequency}} & \multicolumn{3}{c||}{\textbf{Clause Quality}} & \multicolumn{2}{c||}{\textbf{\mcd{} Bursts}} \\ \cline{3-9} 
                               &                                 & A: \fdsc{}                 & B: \fdmc{}                          & C: \albdsc           & D: \albdmc  & E: \avgminalbd{}         & F: \aburst                    & G: \mburst                   \\ \hline
\textbf{SAT}                   & 106                             & 6\%                 & \textbf{10\%}                 & 22.32            & 32.30   & \textbf{18.90}  & 2.69                      & 33.76                    \\ \hline
\textbf{UNSAT}                 & 110                             & 7\%                 & \textbf{12\%}                 & 236.26           & 389.68  & \textbf{80.80}  & 2.70                    & 52.37                    \\ \hline
\textbf{UNSOLVED}              & 184                             & 9\%              & \textbf{16\%}                 & \textbf{72.14}   & 144.75  & 73.38           & 2.60                      & 29.70                    \\ \hline
\textbf{Combined}              & 400                             & 8\%                 & \textbf{13\%}                 & 80.68           &  104.07  & \textbf{60.86}  & 2.65                     & 94.51                    \\ \hline
\end{tabular}
}
\end{table}

\subsection{Distributions of \scd{} and \mcd{} decisions}
We denote \textit{Percentage of Decisions with Single Conflict} and \textit{Percentage of Decisions with Multiple Conflicts} as \fdsc{} and \fdmc{}, respectively. Columns A and B in Table \ref{TabCGD} show the average \fdsc{} and \fdmc{} values for the test instances, under 
SAT, UNSAT and UNSOLVED. Overall, 8\% of all decisions are \scd{}
and 13\% are \mcd{} (see the bottom row). Almost two thirds of all conflict producing decisions are \mcd{}. On average, about 21\% (8\%+13\%) of the decisions are conflict producing. 

This means that on average 79\% of all decisions do not create any conflict.  However, since the \mcd{} decisions produce 2.65 (Column F) conflicts on average, this results in the generation of almost 1 conflict per 2 decisions, on average, which is reflected in the average GLR value of 0.49 for these instances. 

\subsection{Learned Clause Quality in \scd{} and \mcd{} Decisions}
\begin{figure} [h]
        \centering
        \includegraphics[width=\linewidth]{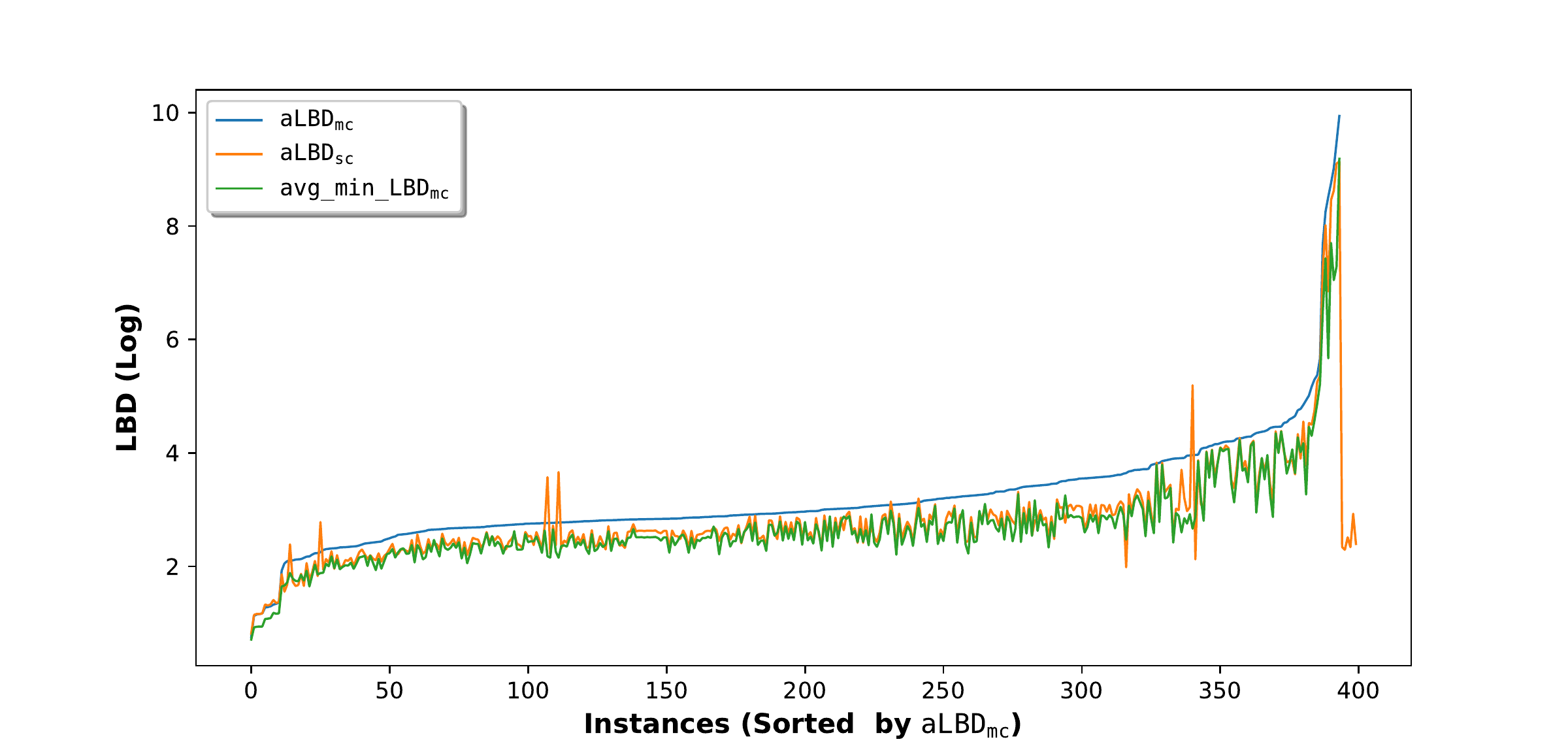}
        \caption{Clause Quality in Conflict Generating Decisions}
        \label{figclsquality}
    \end{figure}
Columns C and D in Table \ref{TabCGD} compare LBD scores when averaged over \scd{} and \mcd{} decisions. On average, \scd{} decisions generate higher quality learned clauses (with lower LBD scores). However, Column E shows that in most cases, the minimum LBD score over the clauses in a single \mcd{} decision is lower on average than for \scd{}. The exception is the UNSOLVED category. Fig. \ref{figclsquality} shows per-instance details of these three measures in log scale. In almost all instances, LBD scores for \mcd{} (blue) are higher than for \scd{} (orange), and minimum \mcd{} LBD (green) is lowest.

To summarize, on average \mcd{} decisions are conflict-inefficient compared to \scd{} decisions. However, on average the best quality learned clause from a \mcd{} decision has better quality than the quality of a \scd{} clause. 
\subsection{Bursts of \mcd{} Decisions}
\begin{figure} [h]
        \centering
        \includegraphics[width=\linewidth]{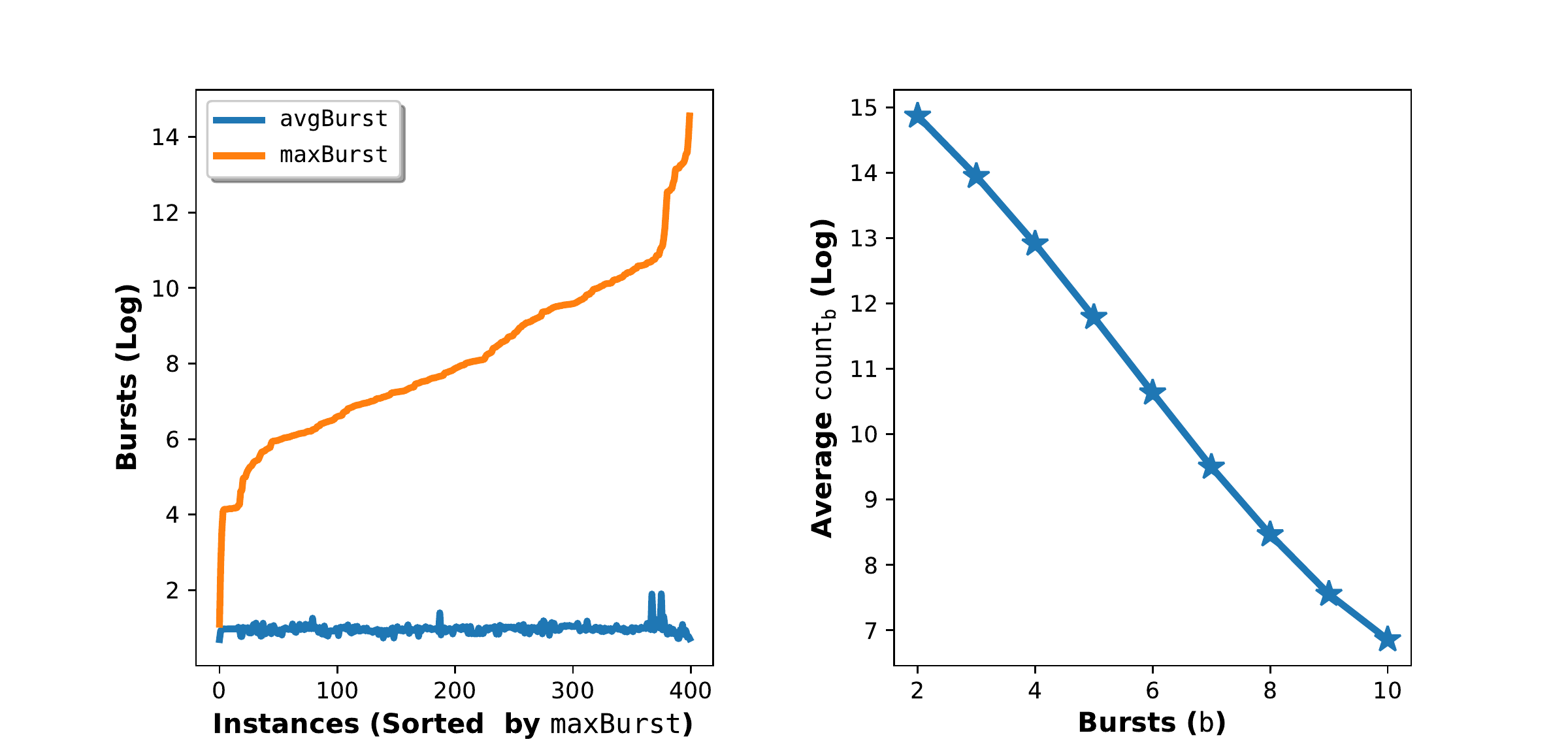}
        \caption{Analysis of Bursts of \mcd{} decisions}
        \label{figburstfreq}
    \end{figure}

Column F in Table \ref{TabCGD} shows the average value of \aburst{} for the test set. On average, the \burst{} of \mcd{} decisions are quite small, about 2.65. However, as shown in column G, the average value of \mburst{} is very high.
The left plot in Fig. \ref{figburstfreq} compares these values for each test instance in log scale. 
In almost all cases \mburst{} (orange) is much larger than the average (blue).  
This indicates that while large bursts of \mcd{} decisions occur, they are rare, as indicated by the average of 2.65. To analyze this in detail, we count the number of \mcd{} decisions for each \burst{} size from 2 to 10.

\medskip
\noindent
{\bf Distribution of \mcd{} Decisions by Burst Size}
Column G of Table \ref{TabCGD} illustrates that \mburst{} can be very large. To simplify our quantitative analysis we focus on counting \mcd{} decisions with bursts up to 10. The plot on the right of Fig. \ref{figburstfreq} shows the average (over the 400 instances in our test set) count (in log scale) of the number of bursts of a given size $b$, with $2 \le b \le 10$. The frequency of bursts decreases exponentially with their size. 

\section{Clause Learning in \mcd{} Decisions}
In this section, we 
establish a structural property of the learned clauses in \mcd{} decisions.

\medskip
\noindent
{\bf Formalization of \mcd{} Decisions} Let $v$ be the decision variable for the \mcd{} decision $\mathcal{M}$ with \burst{} $x \ge 2$.  
At the time when the search reaches the first conflict $C_1$ in $\mathcal{M}$, let $P_0$ be the set of literal assignments that follows from the assignment of $v$. With $1\le i \le x$, let $C_i$ be the $i^{th}$ conflicting clause, from which the clause
$L_i=R_i \lor \{\lnot f_i\}$ is learned. Here, $R_i$ is the reason clause and $f_i$ is the fUIP literal for this $i^{th}$ conflict. After learning $L_i$, and after backtracking, $\lnot f_i$ is the only unassigned literal in $L_i$, and it is immediately unit-propagated from $L_i$. Let $P_i$ be the propagation block that contains literal assignments starting from the assignment of $\lnot f_i$ until the search reaches the conflicting clause $C_{i+1}$.   
Let $\mathcal{L} = (L_1, \dots, L_x)$ be the ordered sequence of $x$ learned clauses in $\mathcal{M}$. 
\begin{figure} [h]
        \centering
        \includegraphics[width=0.8\linewidth]{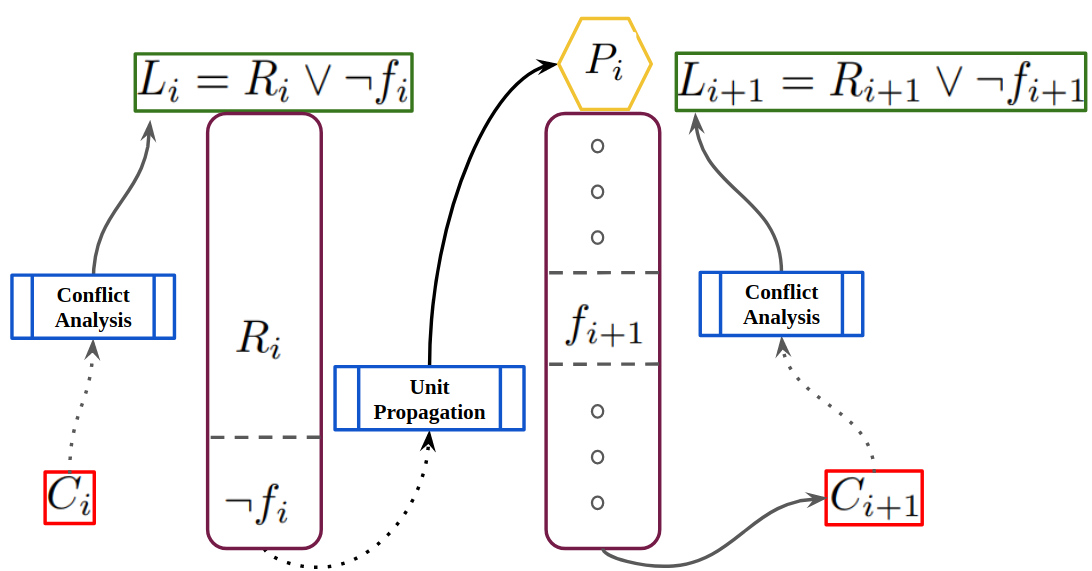}
        \caption{Connection between learned clauses in $\mathcal{M}$. After backtracking, $L_i$ forces $\lnot f_i$, the negated literal of the fUIP for the conflict at $C_i$. This forced assignment creates a block of assignment $P_i$, until the search reaches a conflict at $C_{i+1}$. $P_i$ contains $f_{i+1}$, which is the fUIP of the conflict at $C_{i+1}$. From $C_{i+1}$, Conflict Analysis learns $L_{i+1}$, which contains $\lnot f_{i+1}$. }
        \label{fig_learned_clause_connection}
    \end{figure}

\begin{clm}
\label{proposition1}
$\mathcal{M}$ learns a sequence of clauses $\mathcal{L} = (L_1,  \dots, L_x)$, where a clause $L_i ~(1 \leq i <x)$ implicitly constructs $L_{i+1}$, by implying $f_{i+1}$, the fUIP literal for the $(i+1)^{th}$ conflict, from which $L_{i+1}$ is learned. 
\end{clm}

We justify Claim \ref{proposition1} as follows: 

With $1\le i < x$, after learning the $i^{th}$ clause $L_i = R_i \lor \lnot f_i$ and backtracking to a previous level, the literal $\lnot f_i$ (the negated literal of the fUIP of $i^{th}$ conflict) is forced in $L_i$. This forced assignment creates a propagation block $P_i$ and reaches the conflicting clause $C_{i+1}$.  From $C_{i+1}$ the search learns the next clause $L_{i+1}=R_{i+1} \lor \lnot f_{i+1}$ within the current \mcd{} decision. Clearly, 
the fUIP of $i+1^{th}$ conflict, $ f_{i+1} \in P_i$, as $\lnot f_{i+1} \in L_{i+1}$ is the only literal assigned in the current decision level. Fig. \ref{fig_learned_clause_connection} shows the connection between $L_i$ and $L_{i+1}$.

We have
$(a) (L_{i}=R_i \lor \lnot f_i) \rightarrow  f_{i+1}, \;\;\; (b) f_{i+1} \in P_i , \;\;\; \textnormal{and}\;\;(c) \lnot f_{i+1} \in L_{i+1}$

Hence, under the current partial assignment, the learning of $L_i$ is a \textit{sufficient condition} for the learning of $L_{i+1}$.
Any pair of consecutive clauses $(L_i,  L_{i+1})$ are connected via the pair of assignments $(\lnot f_i, f_{i+1})$, where the first assignment in this pair is the negated literal of the fUIP literal for the $i^{th}$ conflict and the second assignment is the fUIP literal for the $(i+1)^{th}$ conflict. 

Since the argument applies to all $1 \leq i < x$, we have the desired result. 
\section{Proximity between Conflicts Sequences in CDCL}
By Claim \ref{proposition1}, we see that learned clauses in a \mcd{} decision are connected. This indicates that conflicts in a \mcd{} decision are also related, as clauses are learned from conflicts. Here, we first introduce the measure of \confcloseness{} to study proximity between conflict sequences and then present an empirical study to reveal insights on proximity between conflicts sequences in CDCL. 

\subsection{Conflicts Proximity}
The notion of conflicts proximity uses a novel measure called \textit{Literal Block Proximity}, which measures the commonality of literal blocks between a sequence of \textit{reason clauses} over a sequence of conflicts. 
\subsubsection{Literal Block Proximity}  Assume that from a conflicting clause $C$, $L=R\lor {\lnot f}$ is learned, where $R$ is the reason clause for the conflict at $C$ and $f$ is the fUIP literal of current conflict. We define a mapping
$$\ctodl{}:\mathtt{Clause} \longmapsto \mathtt {\{dl_1 \dots dl_n\}}$$
which maps a given reason clause $R$ to the set of distinct decision levels in $R$. Each $\mathtt{dl} \in \ctodl{} (R)$ corresponds to the block of literals $\block_\mathtt{dl}$ in $R$ which were assigned in $\mathtt{dl}$. 

Let $\mathcal{R_{\mathcal{C}}}=(R_1, \dots , R_m)$ be the sequence of reason clauses for the conflicting clauses in $\mathcal{C}=(C_1, \dots C_m)$, where $R_i \in \mathcal{R}$ is the reason clause for the conflict at $C_i \in \mathcal{C}$.
We define the set \textit{Literal Block Proximity} (LBP) for  $\mathcal{R}_\mathcal{C}$, $\lbprc{}$ by
$$\lbprc{} = \ctodl{}(R_1) \cap \dots \cap \ctodl{}(R_m)$$
That is, 
$\lbprc{}$ is the set of decision levels that are common in all clauses in $\mathcal{R}$. Therefore, the assignments in $\mathtt{block_{dl}}$ with $\texttt{dl} \in \lbprc{}$, contribute to the discovery of every conflicting clause in $\mathcal{C}$. 
\begin{examp} 
\label{examp1}
Let $\rc{}= (R_a, R_b)$ be a set of reason clauses for the conflicts at clauses in  $\mathcal{C}= (C_a, C_b)$.
Let $\ctodl(R_a) =$ \{2, 9, 14, 35, 110\} and $\ctodl(R_b)=$ \{9,  10, 11, 35, 98, 110\} be the sets of decision levels in $R_a$ and $R_b$, respectively. Then 
$\lbprc{} = \ctodl(R_a) \cap \ctodl(R_b)$ = \{9, 35,  110\}. The assignments in $\block_{9}$,  $\block_{35}$, and $\block_{110}$ contribute to the generation of conflicts in both $C_a$ and $C_b$. 
\end{examp}
\subsubsection{\textnormal{\confcloseness{}}} 
For a reason clause sequence $\rc=(R_1, \dots , R_m)$, we define the \confcloseness{} $\pfrc{}$, with $0\le \pfrc{} \le 1$ as 

$$\pfrc{} = \frac{\vert \lbprc{} \vert}{\vert \urc \vert}  $$
where $\urc =  \ctodl{}(R_1) \cup \dots \cup \ctodl{}(R_{m}) $ is the set of all literal blocks in $\rc{}$.
In Example \ref{examp1}, $\pfrc=\frac{\vert \lbprc{} \vert}{\vert \urc \vert} = 3/8$. 

Intuitively, for any two given reason clause sequences $\rc{}_p$ and $\rc{}_q$, with $\vert \rc{}_p \vert = \vert \rc{}_q \vert$, if $\proffac_{\rc{}_p} > \proffac_{\rc{}_q}$, then the conflicts associated with the reason clauses in $\rc{}_p$ are more closely related to each other than conflicts associated with the reason clauses in $\rc{}_q$. 
If $\proffac_{\rc{}_p} > \proffac_{\rc{}_q}$, then we call the conflicts generated over the clauses in $\mathcal{C}_p$ \textit{more} \proximated{} than the conflicts generated over the clauses in $\mathcal{C}_q$. 

\begin{examp}
Let $\rc{}_p=(R_{p1}, R_{p2})$ and $\rc{}_q=(R_{q1}, R_{q2} )$ be two sequences of reason clauses for conflicts generated at the conflicting clauses in $\mathcal{C}_p=(C_{p1}, C_{p2})$ and $\mathcal{C}_q=(C_{q1}, C_{q2})$, respectively. Let $\proffac_{\rc{}_p}=0.65$ and $\proffac_{\rc{}_q}=0.35$. Then conflicts generated over the conflicting clauses in $\mathcal{C}_p$ are more \proximated{} than conflicts generated over the conflicting clauses in $\mathcal{C}_q$.
\end{examp}

We now study proximity of conflicts in CDCL under \confcloseness{}.
\subsection{Proximity of Conflicts over \scd{} and \mcd{} Decisions}
 While the learned clauses in a \mcd{} decision are connected, each learned clause in a \scd{} decision is learned in isolation. Based on this observation, we propose the following hypothesis: 
 \begin{hyp}
 On average, conflicts in a \mcd{} decision with burst $x$ are more \proximated{} than conflicts which are generated in the last $x$ \scd{} decisions. 
 \end{hyp}
\vspace{-0.2cm}

 We support this hypothesis by comparing the \confcloseness{} of reason clauses over \mcd{} and \scd{} decisions. 
 \vspace{-0.4cm}
  \begin{figure} [h]
        \centering
        \includegraphics[width=\linewidth]{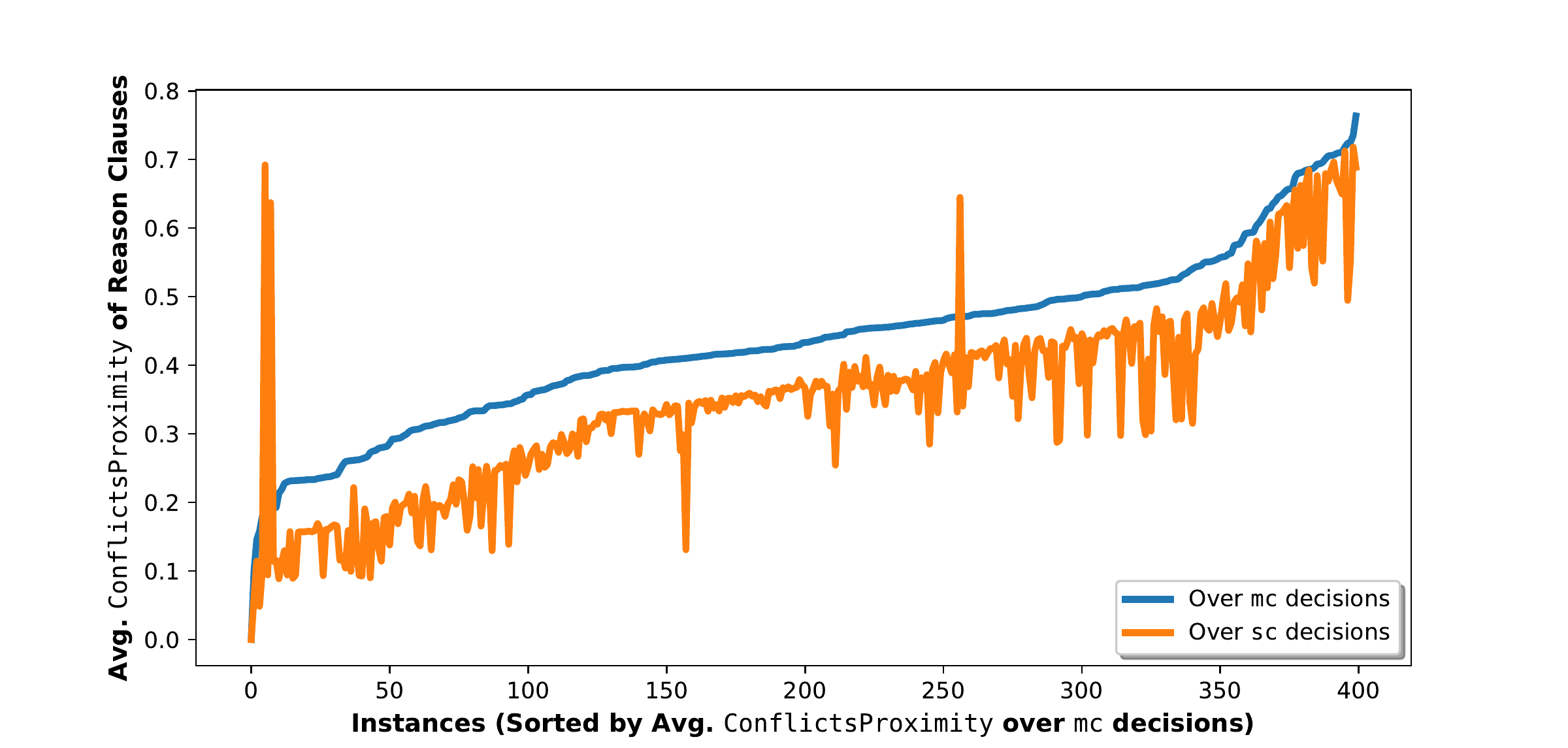}
        \caption{Comparison of average \confcloseness{} of Reason Clauses over \mcd{} and \scd{} Decisions}
        \label{lbp_mcsc}
    \end{figure}
\vspace{-0.3cm}


\medskip
\noindent
{\bf Experiment} We have performed an experiment with 400 instances from \satcomp{} with \mpldl{} with a time-limit of 5,000 seconds. For each run of an instance, whenever the search finds a \mcd{} decision with burst $x$, we compute the LBP and \confcloseness{} for the reason clauses (i) for these $x$ conflicts  and (ii) for the last $x$ conflicts in \scd{} decisions. For this experiment, we collect data for bursts $x\le 10$. For a run with an instance, we compute the average of \confcloseness{} of reason clauses separately over \mcd{} and \scd{} decisions.

 Fig. \ref{lbp_mcsc} shows that average \confcloseness{} for the reason clauses over \mcd{} decisions (blue lines, average is 0.43) are higher than \confcloseness{} of reason clauses over \scd{} decisions (orange line, average is 0.34)  for almost all instances. This validates our hypothesis that conflicts over $\mcd{}$ decisions are more \proximated{} than conflicts over $\scd{}$ decisions. 

\section{The Common Reason Variable Reduction Strategy}\label{crvr}
\subsection{Common Reason Decision Variables} 
Assume that a \mcd{} decision $\mathcal{M}$ finds $x \ge 2$ consecutive conflicts within its decision. Let $\mathcal{R}=(R_1, \dots,  R_x)$ be the sequence of reason clauses for these $x$ conflicts. $\mathbf{LBP_{\mathcal{R}}}$ is the set of common decision levels over $\mathcal{R}$. For each decision level $\mathtt{dl} \in \mathbf{LBP_{\mathcal{R}}}$, we call $\mathtt{dl}$, a \textit{common reason decision level} and the decision variable $v$ at $\mathtt{dl}$, a \textit{common reason decision variable (CRV)} for $\mathcal{M}$. If $\vert \mathbf{LBP_{\mathcal{R}}}\vert=z$, then there are $z$ CRVs in $\mathcal{M}$. The CRVs in $\mathcal{M}$ are the decision variables from previous decision levels, which contributed to the generation of all the conflicts in $\mathcal{M}$. 

\subsection{Poor \mcd{} Decisions}

Recall that in Section \ref{section_mc_sc_study} (Fig. \ref{figclsquality}),
we observed that on average, \mcd{} decisions (blue line) produce lower quality clauses than \scd{} decisions (orange line). However, the best quality clause (green line) in a \mcd{} decision has better average quality than other learned clauses.  
However, in a  \poortilted{} \mcd{} decision its best quality learned clause is worse than the average quality. A \mcd{} decision $\mathcal{M}$ is \poortilted{},
\begin{itemize}
    \item if the quality of the best learned clause in $\mathcal{M}$ is lower than a dynamically computed threshold $\theta$, the average quality of the last $k$ learned clauses.
\end{itemize}
\vspace{-0.3cm}
\begin{table}[h]
\centering
\caption{The \crvr{} Decision Strategy}
\label{TabCRVR}
\resizebox{\columnwidth}{!}{
\begin{tabular}{l@{\hspace{0.2cm}}|@{\hspace{0.2cm}}l}
\multicolumn{1}{c}{\trackcrv{}}                            & \multicolumn{1}{c}{\reducecrv{}}                                   \\
\begin{tabular}[c]{@{}l@{}}
1. $\theta \leftarrow aLBD(k)$
\\
2. if ($\minalbd_{\mathcal{M}} \; > \;\theta)$\\
$\;\;\;$a. $\mathcal{R} \leftarrow reason\_clauses\_in\_\mathcal{M}()$\\$\;\;\;$b. for each $\mathtt{dl} \in \mathbf{LBP_\mathcal{R}}$\\    
$\;\;\;\;\;\;$i. $\mathtt{v}\leftarrow dvar(\mathtt{dl})$ \\ $\;\;\;\;\;\;$ii. \poortilted{}\texttt{\_crv[v]} $\leftarrow$ true
\end{tabular} & \begin{tabular}[c]{@{}l@{}}1. \texttt{selected} $\leftarrow$ false\\ 2. while (\textit{not} \texttt{selected})\\   $\;\;\;\;$a. $\mathtt{y}\leftarrow$ \textit{select\_next\_free\_var()}\\   $\;\;\;\;$b. if (\poortilted{}\texttt{\_crv[y]})\\
$\;\;\;\;\;\;\;$ i. \texttt{activity[y]} $\leftarrow$ \texttt{activity[y] * (1-Q)}\\ $\;\;\;\;\;\;\;\;$ii. \poortilted{}\texttt{\_crv[y]} $\leftarrow$ false\\ $\;\;\;\;\;\;\;\;$iii. \textit{reorder()}\\ $\;\;\;\;\;$c. else \\ $\;\;\;\;\;\;\;\;$ i. \texttt{selected}$\leftarrow$true\end{tabular}
\end{tabular}
}
\end{table}
 \vspace{-0.5cm}
\subsection{The \crvr{} Decision Strategy}
We summarize the previous two subsections as follows: 
\begin{itemize}
    \item Conflicts in a \poortilted{} \mcd{} decision are not likely to be helpful, as the quality of its best learned clause is lower than the recent search average.
    \item The CRVs in a \poortilted{} \mcd{} decision combinedly contribute to the generation of these conflicts. 
\end{itemize} 
Does suppression of such CRVs for the future decisions help the search to achieve better efficiency? We address this question by designing a decision strategy named \textit{common reason variable score reduction} (\crvr{}), which can be integrated with any activity based variable selection decision heuristics, such as VSIDS and LRB. The high-level idea of \crvr{} is  as follows: Once a \poortilted{} \mcd{} decision is detected,
\crvr{} (i) finds the CRVs for that poor \mcd{} and (ii) marks those CRVs as \poortilted{} {CRVs}, and (iii) then reduces the activity scores of those \poortilted{} {CRVs} for future decisions. \crvr{} consists of the two procedures \trackcrv{} and \reducecrv{}, whose pseudo-codes are shown in Table \ref{TabCRVR}.

\bigskip
\noindent
{\bf \trackcrv{}} This procedure is invoked at the end of an \mcd{} decision $\mathcal{M}$ and just before 
the next decision. It computes a dynamic conflict quality threshold $\theta$, the average LBD score of last $k$ learned clauses. Then it determines if $\mathcal{M}$ is \poortilted{}. by comparing \minalbd{}$_\mathcal{M}$
with $\theta$.  In this case,  \trackcrv{}
obtains the sequence of reason clauses $\mathcal{R}$ in $\mathcal{M}$ and computes $\mathbf{LBP_{\mathcal{R}}}$. For each decision level $\mathtt{dl} \in \mathbf{LBP_{\mathcal{R}}}$, any decision variable $\mathtt{v}$ at $\mathtt{dl}$ is marked as \poortilted{}.

\bigskip
\noindent
{\bf \reducecrv{}} is shown in the right side of Table \ref{TabCRVR}. This procedure modifies a typical CDCL decision routine to lazily reduce the activity score of \poortilted{} CRVs. It employs a while loop until a variable is \texttt{selected}, where in each iteration of the loop, it performs the following operations: (i) obtains a free variable \texttt{y}, where \texttt{y} is the free variable with largest activity score. (ii) checks if \texttt{y} is marked as \poortilted{}. If \texttt{y} is \poortilted{}, 
then it computes a fraction of \texttt{activity[y]}, the current activity score of \texttt{y}, by multiplying it with \texttt{(1-Q)}, where \texttt{Q} is a user defined parameter with $0 < \mathtt{Q} < 1$ . This fraction, which is lower than \texttt{activity[y]}, becomes the new activity score of \texttt{y}. (iii) 
it unmarks \texttt{y} as \poortilted{} and performs a reordering of the variables, which reorders the variables  by their activity scores.
The reduction of the activity score of \texttt{y},  followed by reordering, decreases the selection priority of \texttt{y}.

 \section{Experimental Evaluation}
 \subsection{Implementation}
 We implemented \crvr{} in three state-of-the-art baseline solvers \mpldl{}, \kissat{} and \kissatdef{}. We call the extended solvers \mpldlext{}, \textbf{\kissatext{}}, and \kissatdefext{}, respectively.
 The solver \mpldl{} employs a combination of the decision heuristics DIST \cite{dist}, VSIDS \cite{Chaff} and LRB \cite{LiangGPCSAT16}, which are activated at different phases of the search, whereas  \kissat{} and \kissatdef{} use  VSIDS and Variable Move to Front (VMTF) \cite{RyanMS} alternately during the search. 
 
The heuristics DIST, VSIDS and LRB share similar computational structures. All three heuristics maintain an \textit{activity score} for each variable. Whenever a variable involves in a conflict, its activity score is increased. In contrast, VMTF maintains a queue of variables, where a subset of variables appearing in a learned clause are moved to the front of that queue in an arbitrary order.

\crvr{} is designed to be employed on top of activity-based decision heuristics. Hence in \kissatext{} and \kissatdefext{}, we employ \crvr{} only when VSDIS is active.
 
 In all of our extended solvers, we use the following parameter values: a length of window of recent conflicts  $k =50$ and an activity score reduction factor \texttt{Q} $=0.1$. Source code of our \crvr{} extensions are available at \cite{srcCRVR}. 
  
 \begin{table}[h]
\centering
\caption{Comparison between 3 baselines and their \texttt{\crvr{}} extensions}
\label{TabPerformacneComparison}
\resizebox{0.8\columnwidth}{!}{
\begin{tabular}{|l|l|l|l|l|}
\hline
\multicolumn{1}{|c|}{\textbf{Systems}} & \textbf{SAT}                & \textbf{UNSAT}        & \textbf{Combinded}         & \textbf{PAR-2}          \\ \hline
\mpldl                         & 106                & \textbf{110} & 216               & 2065          \\ \hline
\mpldlext            & \textbf{116 (+10)} & 107 (-3)     & \textbf{223 (+7)} & \textbf{2001} \\ \hline \hline
\kissat                        & 148                & \textbf{118} & \textbf{266}      & \textbf{1552} \\ \hline
\kissatext               & \textbf{150 (+2)}  & 114 (-4)     & 264 (-2)          & 1565          \\ \hline \hline
\kissatdef                     & 134                & \textbf{126} & 260               & 1624          \\ \hline
\kissatdefext            & \textbf{139 (+5)}  & 125 (-1)     & \textbf{264 (+4)} & \textbf{1588} \\ \hline
\end{tabular}
}
\end{table}

\subsection{Experiments and Results}
We conduct experiments with the same set of 400 instances with a 5,000 seconds timeout per instance. We compare the \crvr{} extensions and their counterpart baselines in terms of number of solved instances, 
solving time and PAR-2 score.\footnote{PAR-2 score is defined as the sum of all runtimes for solved instances + 2*timeout for unsolved instances  \cite{par2}. Lower scores are better.}

Table \ref{TabPerformacneComparison} compares \mpldlext{}, \kissatext{}, and \kissatdefext{} with their baselines. All of these extensions show performance gains on SAT instances, but lose on UNSAT instances. 
The strongest gain is for \mpldlext{}, which
 solves 10 additional SAT instances, but solves 3 less UNSAT instances, achieves an overall gain of 7 instances compared to its baseline. \kissatext{} solves 2 more SAT instances, but looses 4 UNSAT instances, with an overall loss of 2 instances compared to its baseline. The third extension \kissatdef{} solves 5 more SAT instances, but solves 1 less UNSAT instance than its baseline. Overall, \kissatdefext{} solves 4 more instances than \kissatdef{}. 
\vspace{-0.3cm}
 \begin{figure} [h]
        \centering
        \includegraphics[width=\linewidth]{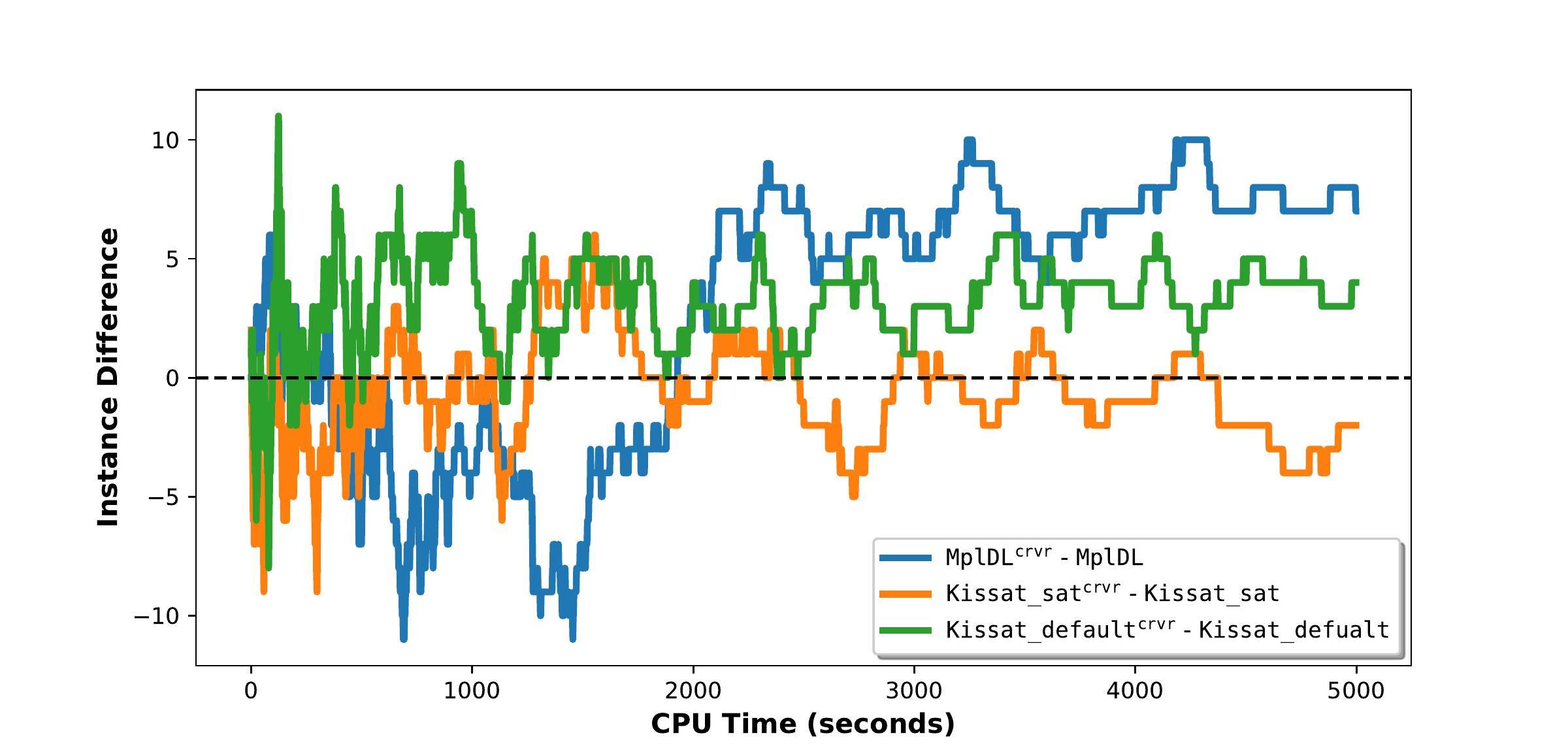}
        \caption{Solve time comparisons: For any point above 0 in the vertical axis, our extensions solve more instances than their baselines at the time point in the horizontal axis.}
        \label{figsolvetime}
    \end{figure}
 The PAR-2 results are consistent with the solution count results. While \kissatext{} has a slight increase (a 0.8\% increase) in its PAR-2 score compared to \kissat{}, both \mpldlext{} and \kissatdefext{} have significantly lower PAR-2 scores (3.19\% and 2.28\%  of reductions, respectively) compared to their baselines, which reflects overall better performance of these two systems compared to their baselines.  
 
 Fig. \ref{figsolvetime} compares the relative solving speed of  \mpldlext{} (blue), \kissatext{} (orange), and \kissatdefext{} (green) against their baselines by plotting the difference in the number of instances solved as a function of time. If that difference is above 0, then it indicates that an extended solver solves more instances than the baseline at this time point.

\mpldlext{} (blue) performs slightly worse than \mpldl{} early on, but beats the baseline consistently after 1,900 seconds. Compared to \kissatdef{}, \kissatdefext{} (green) is ahead of \kissatdef{} at most time-points. \kissatext{} (orange) is behind  \kissat{} at most of the time-points. 

Overall, compared to their baselines, our extensions perform better on SAT instances, but loose a small number of UNSAT instances. 
\section{Detailed Performance Analysis of \crvr{}} 
For a run of a given solver, the metric GLR measures the overall conflict generation rate of the search, average LBD (aLBD) measures the average quality of the learned clauses and G2L measures the fraction of learned clauses which are glue. All these measures correlate well with solving efficiency \cite{CP19,Liang2017}.
 Here, we present an analysis that relates the performance of \crvr{} with these three metrics. We consider two subsets of instances, where \mpldl{} and \mpldlext{} show opposite strengths:
 
\begin{itemize} 
\item $\excmpldl$: 12 instances which are solved by \mpldl{}, but not by \mpldlext{}.
 
\item $\excmpldlext{}$: 19 instances which are solved by \mpldlext{}, but not by \mpldl{}.

\end{itemize}
\vspace{-0.3cm}
  \begin{table}[h]
\centering
\caption{Relating solving efficiency with three metrics average GLR, average LBD and average G2L. More efficient branching heuristics tend to achieve higher average GLR \cite{Liang2017}, higher average G2L \cite{CP19} and lower average aLBD \cite{Liang2017} than less efficient ones.}
\label{TabConfMetrics}
\resizebox{\columnwidth}{!}{
\begin{tabular}{|l|l|l|l|l|l|l|l|}
\hline
\multicolumn{1}{|c|}{\multirow{2}{*}{\textbf{Instance Sets}}} & \multirow{2}{*}{\textbf{Count (SAT+UNSAT)}} & \multicolumn{2}{l|}{\textbf{average GLR}}                          & \multicolumn{2}{l|}{\textbf{average aLBD}}                         & \multicolumn{2}{l|}{\textbf{average  G2L}}                         \\ \cline{3-8} 
\multicolumn{1}{|c|}{}                                        &                                             & \textbf{\mpldl} & \textbf{\mpldlext} & \textbf{\mpldl} & \textbf{\mpldlext} & \textbf{\mpldl} & \textbf{\mpldlext} \\ \hline
\textbf{$\excmpldlext{}$}                                        & 19 (18+1)                                   & \textbf{0.58}                  & 53                                & 4399.91                        & \textbf{147.69}                   & 0.003                          & \textbf{0.018}                    \\ \hline
\textbf{$\excmpldl{}$}                                           & 12 (8+4)                                    & 0.56                           & 0.56                              & 18.83                          & \textbf{18.53}                    & \textbf{0.024}                 & 0.023                              \\ \hline
\end{tabular}
}
\end{table}
\vspace{-0.1cm}

The 19 instances in $\excmpldlext{}$ (first row of Table \ref{TabConfMetrics}) are solved exclusively by \mpldlext{}. For this subset, \mpldlext{} learns clauses at a slightly lower rate. However, for $\excmpldlext{}$, the average aLBD (resp. average G2L) is significantly lower (resp. higher) with \mpldlext{}. \crvr{} helps to (i) learn higher quality clauses, and (ii) learn more glue clauses relative to the number of clauses for the subset of instances in $\excmpldlext{}$, for which \mpldlext{} is efficient. 

18 of the 19 instances in $\excmpldlext{}$ are SAT. The learning of significantly better quality of clauses with $\mpldlext{}$ for these SAT instances may just explain the good performance of \crvr{} on SAT instances. 

For the 12 instances in $\excmpldl{}$ (second row of Table \ref{TabConfMetrics}), \mpldl{} learns clauses at the same rate, but learns clauses which are of slightly lower quality than the clauses learned by \mpldlext{}. However, for this set, the average G2L value is slightly higher with \mpldl{} than \mpldlext{}. This could explain the better performance of \mpldl{} for this subset.

 \section{Related Work}
  Audemard and Simon \cite{extremeCases} briefly studied \textit{decisions with successive conflicts}, which we refer to as \mcd{} decisions in this paper. They studied number of successive conflicts in the CDCL solver Glucose on a fixed set of instances. In the current paper, we present a more formal and in-depth study of \mcd{} decisions. The authors of \cite{Liang2017} relate conflict generation propensity and learned clause quality with the efficiency of several decision heuristics. In contrast, we study and compare the conflict quality of two types of conflict producing decisions for CDCL. Conflicts generation pattern in CDCL is studied in \cite{expSATAAAI2020} showing that CDCL typically alternates between bursts and depression phases of conflict generation. While that work presented an in-depth study of the conflict depression phases in CDCL, here we study the the conflict bursts phases, which are opposite of conflict depression phases. Chowdhury et al. \cite{CP19} studied conflict efficiency of decisions with two types of variables: those that appear in the glue clauses and those that do not. In the current paper, we compare the conflict efficiency of conflict producing decisions.  
 \section{Conclusions and Future Work} We present a characterization of \scd{} and \mcd{} decisions in terms of average learned clause quality that each type produces. Then we analyze how \mcd{} decisions with different bursts are distributed in CDCL search. Our theoretical analysis shows that learned clauses in a \mcd{} are connected, indicating that conflicts that occur in a \mcd{} decision are related to each other. We introduced a measure named \confcloseness{} that enables the study of proximity of conflicts in a given sequence of conflicts. Our empirical analysis shows that conflicts in \mcd{} decisions are more \proximated{} than conflicts in \scd{} decisions. Finally, we formulated a novel CDCL strategy \crvr{} that reduces the activity score of some variables that appear in the clauses learned over \mcd{} decisions. Our empirical evaluation with three modern CDCL SAT solvers shows the effectiveness of \crvr{} for the SAT instances from \satcomp{}. 
 
 In the future, we intend to pursue the following research questions: 
\begin{itemize}
\item Kissat solvers and their predecessors, such as CaDiCaL, employ VMTF as one of their decision heuristics. How to extend VMTF with \crvr{} is an interesting question that we plan to pursue in future. 
 
\item Currently, the user defined parameters $k$ and $\texttt{Q}$ in  \crvr{} are set to fixed values. How to adapt them dynamically during the search? We hope that a dynamic strategy to adapt these parameters will improve the performance of \crvr{}, specially over the UNSAT instances. 
\end{itemize}
\bibliographystyle{plain}
\bibliography{paper}
\end{document}